\documentclass[10pt,twocolumn,letterpaper]{article}

\usepackage{cvpr}              

\usepackage{graphicx}
\usepackage{amsmath}
\usepackage{amssymb}
\usepackage{booktabs}
\usepackage{multirow}
\usepackage[accsupp]{axessibility} 
\usepackage[breaklinks=true,
            bookmarks=false,
            colorlinks,
            linkcolor=red,       
            anchorcolor=blue,  
            citecolor=green, 
            ]{hyperref}

\usepackage[capitalize]{cleveref}
\crefname{section}{Sec.}{Secs.}
\Crefname{section}{Section}{Sections}
\Crefname{table}{Table}{Tables}
\crefname{table}{Tab.}{Tabs.}

\def\cvprPaperID{2882} 
\def\confName{CVPR}
\def\confYear{2022}

\begin{document}


\title{\textbf{HerosNet:} \textbf{H}yperspectral \textbf{E}xplicable \textbf{R}econstruction and \textbf{O}ptimal \textbf{S}ampling Deep \textbf{Net}work for Snapshot Compressive Imaging}

\author{
Xuanyu Zhang$^{1}$, Yongbing Zhang$^{2}$, Ruiqin Xiong$^{3}$, Qilin Sun$^{4}$, Jian Zhang$^{1,5}$\\
$^{1}$Peking University Shenzhen Graduate School, Shenzhen, China\\
$^{2}$Harbin Institute of Technology, Shenzhen, China $\quad^{3}$Peking University, Beijing, China\\
$^{4}$The Chinese University of Hong Kong, Shenzhen, China $\quad^{5}$Peng Cheng Laboratory, Shenzhen, China\\
}

\maketitle 
\let\thefootnote\relax\footnotetext{This work was supported in part by Shenzhen Fundamental Research Program (No.GXWD20201231165807007-20200807164903001) and National Natural Science Foundation of China (61902009). (\textit{Corresponding author: Jian Zhang.})}
\begin{abstract}
Hyperspectral imaging is an essential imaging modality for a wide range of applications, especially in remote sensing, agriculture, and medicine. Inspired by existing hyperspectral cameras that are either slow, expensive, or bulky, reconstructing hyperspectral images (HSIs) from a low-budget snapshot measurement has drawn wide attention. By mapping a truncated numerical optimization algorithm into a network with a fixed number of phases, recent deep unfolding networks (DUNs) for spectral snapshot compressive sensing (SCI) have achieved remarkable success. However, DUNs are far from reaching the scope of industrial applications limited by the lack of cross-phase feature interaction and adaptive parameter adjustment. In this paper, we propose a novel \textbf{H}yperspectral \textbf{E}xplicable \textbf{R}econstruction and \textbf{O}ptimal \textbf{S}ampling deep \textbf{Net}work for SCI, dubbed \textbf{HerosNet}, which includes several phases under the ISTA-unfolding framework. Each phase can flexibly simulate the sensing matrix and contextually adjust the step size in the gradient descent step, and hierarchically fuse and interact the hidden states of previous phases to effectively recover current HSI frames in the proximal mapping step. Simultaneously, a hardware-friendly optimal binary mask is learned end-to-end to further improve the reconstruction performance.  Finally, our HerosNet is validated to outperform the state-of-the-art methods on both simulation and real datasets by large margins. The source code is available at \href{https://github.com/jianzhangcs/HerosNet} {https://github.com/jianzhangcs/HerosNet}.

\end{abstract}

\section{Introduction}
\label{sec:intro}

With the development of artificial intelligence and robotics, the demand for capturing and sensing hyperspectral images has dramatically increased in recent years \cite{yuan2021snapshot,arce2013compressive,cao2016computational,he2021fast}. Based on the traditional compressive sensing (CS) \cite{zhao2016video, zhang2014image}, spectral snapshot compressive sensing (SCI) system aims to record 3D scenes via a 2D detector. It has the advantages of low bandwidth, low cost, and high data throughput, which has played an increasingly pivotal role in a wide range of applications, such as remote sensing, object detection, super-resolution, and medical diagnosis \cite{zhao2015sub,ojha2015spectral,melgani2004classification,lei2019spectral,xie2019structure,ding2020hyperspectral,wei2019medical,akbari2010detection}. In this paper, we focus on a typical imaging system named coded aperture snapshot spectral imager (CASSI) \cite{wagadarikar2009video,gehm2007single,meng2020snapshot}, which modulates spectral frames via a coded aperture (i.e. physical mask) and shifts them across the spectral dimension via a disperser. 
\begin{figure}[t]
	\centering
	\includegraphics[width=1\linewidth]{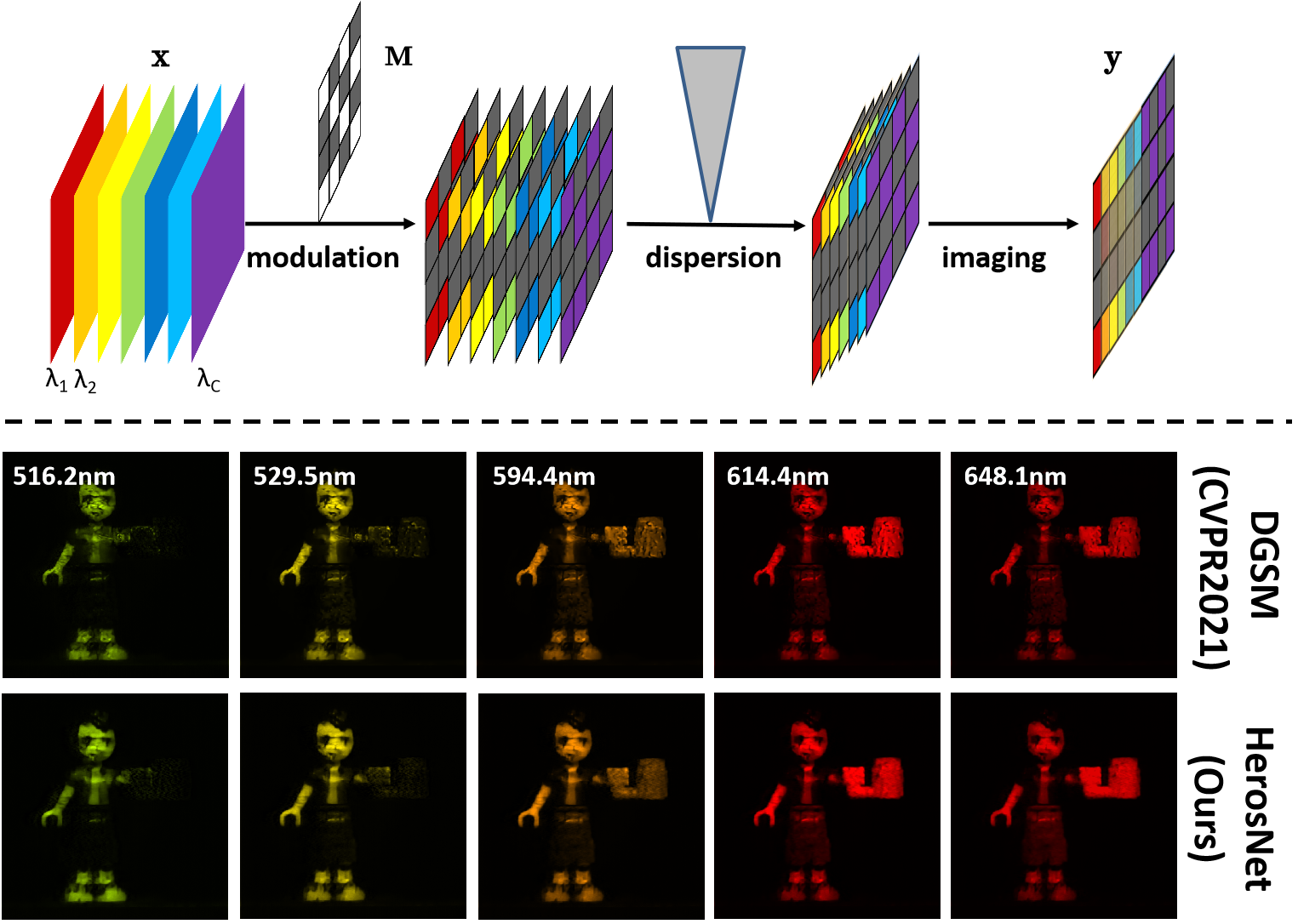}
	\caption{The schematic of CASSI system and some visual results of the proposed HerosNet and DGSM \cite{huang2021deep} on the real dataset. Our reconstructed HSIs have clearer edges and more detailed textures, while the results of DGSM have more noise and artifacts.}
	\label{intro}
\end{figure}

In the past few years, there have been a great amount of reconstruction methods for HSIs compressed by CASSI, including model-based methods and deep learning-based methods. Traditional model-based methods \cite{tan2015compressive, wang2021tensor, bioucas2007new, yuan2016generalized, lin2014spatial, yang2014video, yang2014compressive, liu2018rank} often search for the optimal solution iteratively and refine the results to the desired signal domain via the image priors. Although these methods are highly interpretable, they are limited by the hand-crafted priors and slow reconstruction speed. Owing to the development of deep learning, several learning-based methods \cite{zhao2021lstm,xiong2017hscnn,shi2018hscnn+,miao2019net,meng2020end,zheng2021deep,meng2021self,meng2021perception, Yorimoto_2021_ICCV} have been used to reconstruct HSIs, which directly learn an end-to-end inverse mapping from the 2D measurement to the 3D hyperspectral cube. Compared to model-based methods, they tend to drastically reduce time complexity and achieve better performance. However, they do not explicitly embody the system imaging model and are just trained as a black box.

Most recently, some researchers have introduced the DUNs to the HSI reconstruction task, which merge the advantages of both model-based and learning-based methods \cite{wang2019hyperspectral,wang2020dnu,meng2020gap,ma2019deep,huang2021deep,zhang2021learning}. DUNs perform the iterative process via a gradient descent module and refine the intermediate result via a deep prior module. Although existing DUNs alleviate some drawbacks of model-based and learning-based methods to some extent, there are still several bottlenecks to be solved. Firstly, how to effectively interact and fuse the features between phases is the key to the enhancement of the reconstruction quality. Most existing DUNs do not establish the connections between the proceeding and following phases. As the number of phases increases, beneficial information tends to be lost in the process of information transmission. Furthermore, inspired by the memory mechanism in the CS \cite{song2021madun}, the hidden states of the previous phases can provide complementary information for the computation of the current phase. Therefore, it is essential to introduce a feature interaction mechanism between different phases to obtain enhanced feature representation. Secondly, how to dynamically learn the parameters in the gradient descent module has been ignored in the past. Previous researchers usually treat these parameters as fixed constants to compress HSIs. However, the fixed parameters can not be adjusted adaptively and contextually in different scenes, which will lead to sub-optimal reconstruction and restrict the flexibility of DUNs. Thirdly, targeted at HSIs, the existing DUN does not combine mask optimization and image reconstruction into a united framework, which can not retain the structure and information of HSIs completely. 

Inspired by the ISTA, a novel \textbf{H}yperspectral \textbf{E}xplicable \textbf{R}econsturction and \textbf{O}ptimal \textbf{S}ampling deep \textbf{Net}work for SCI, dubbed \textbf{HerosNet}, is proposed for joint mask optimization and HSI reconstruction. Particularly, the network consists of a sampling subnet, an initialization subnet, and a recovery subnet. Fueled by the success of ISTA-Net and its variant \cite{zhang2018ista, you2021ista, wu2021dense}, the recovery phase updates the current estimate via a dynamic gradient descent module (DGDM) and refine the rough estimate via a hierarchical feature interaction module (HFIM). Since each recovery phase corresponds to an ISTA iteration and all the parameters are learned end-to-end, the network enjoys the merits of high-quality reconstruction with strong interpretability. For example, in Fig.~\ref{intro}, the results of the proposed HerosNet have clearer detailed textures than those of DGSM \cite{huang2021deep}. Overall, our contributions are summarized as follows.
\begin{itemize}
	\item A novel ISTA inspired deep unfolding network, dubbed HerosNet, is proposed for jointly learning binary optimal masks and recovering high-quality HSIs.
	
	\item A dynamic gradient descent module (DGDM) is introduced to flexibly simulate the sensing matrix and contextually adjust the step size in gradient descent step. 
	
	\item A hierarchical feature interaction module (HFIM) is designed, which fuses and interacts the hidden states of previous phases to recover the HSI frames of the current phase in proximal mapping step.	
	
	\item Our HerosNet outperforms the state-of-the-art methods on simulation and real datasets by large margins.
\end{itemize}

\begin{figure*}[]
	\centering
	\includegraphics[width=1\linewidth]{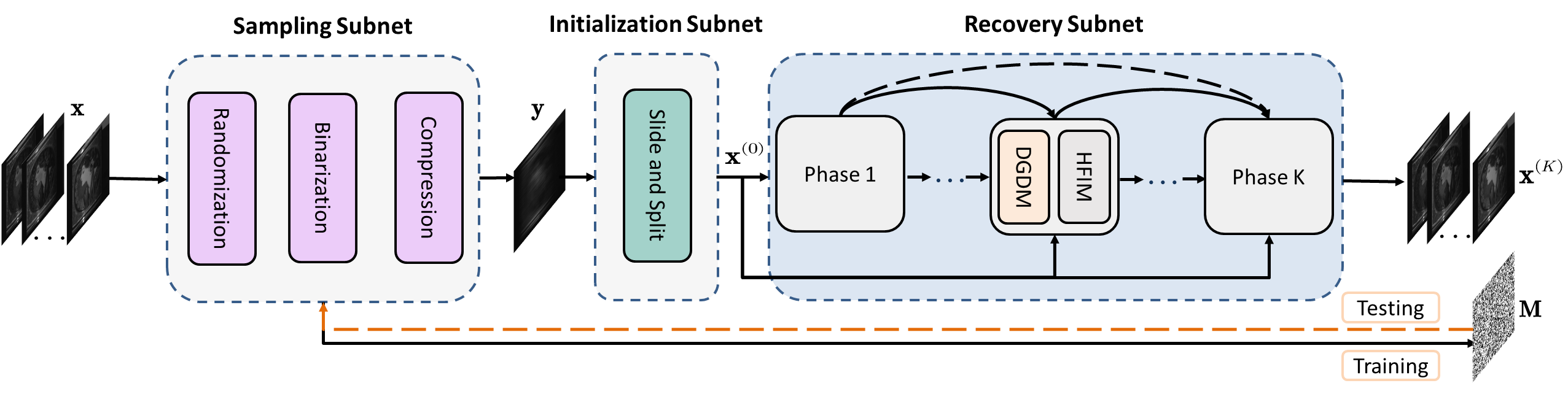}
	\caption{Illustration of the proposed HerosNet, including sampling subnet, initialization subnet and recovery subnet. The recovery subnet consists of $K$ phases. Each phase is composed of the dynamic gradient descent module (DGDM) and the hierarchical feature interaction module (HFIM). In the training procedure, the HerosNet takes the 3D hyperspectral cube $\mathbf{x}$ as input to obtain the compressed measurement $\mathbf{y}$, and generates the reconstructed HSI frames $\mathbf{x}^{(K)}$ and the optimal mask $\mathbf{M}$. In the testing procedure, the HerosNet compresses the hyperspectral cube $\mathbf{x}$ via the optimized binary mask $\mathbf{M}$ and reconstructs HSI frames $\mathbf{x}^{(K)}$.}
	\label{net}
\end{figure*}

\section{Related Works}
\subsection{HSI Reconstruction Algorithms}
\noindent \textbf{Model-based Methods:} The model-based methods employ the regularization term inspired by the image prior to solve the ill-posed inverse problem iteratively. In \cite{bioucas2007new}, a two-step iterative shrinkage/thresholding (TwIST) algorithm was designed to keep the good reconstruction performance and improve the speed of convergence. In \cite{yuan2016generalized}, the total variation optimization was applied in the HSI reconstruction and the generalized alternating projection (GAP) algorithm was utilized to solve the optimization problem. In \cite{lin2014spatial}, an overcomplete dictionary was learned to exploit the sparsity representation and reconstruct the HSIs. In \cite{yang2014video,yang2014compressive}, the HSI reconstruction task was treated as the maximum likelihood estimation and the Gaussian mixture model (GMM) was introduced to model the data distribution of the HSIs. Recently, in \cite{liu2018rank}, the non-local self-similarity of HSIs and rank minimization strategy were incorporated into the framework of alternating direction method of multipliers (ADMM), which has achieved the best performance among the traditional model-based methods. Although these methods produce decent results in specific applications, it is difficult to design hand-crafted priors suitable for all scenes.

\noindent \textbf{Deep Learning-based Methods:} Relying on the powerful representation ability of deep networks, the learning-based HSI reconstruction methods have attracted more and more attention. The learning-based methods are divided into two categories generally according to whether they are unfolded from the optimization process. Among the methods without deep unfolding, the end-to-end deep networks tend to directly learn a non-linear mapping from the 2D measurement to the 3D hyperspectral cube. For instance, Miao \emph{et al.} \cite{miao2019net} introduced the dual-stage generative model to extract both spectral and spatial information. Meng \emph{et al.} \cite{meng2020end} embedded three self-attention modules into the U-Net backbone, thus achieving high-quality and real-time reconstruction. Apart from the end-to-end networks, the plug-and-play (PnP) framework \cite{zheng2021deep,yuan2020plug} incorporated the pre-trained deep denoisers into the optimization process and effectively projected the image signal to the desired domain. Most recently, deep image prior was integrated with the PnP regime to construct an untrained self-supervised network \cite{meng2021self}. Although these methods perform a certain role in reconstructing HSIs, they all face some unavoidable challenges. For instance, the end-to-end deep networks are lack of interpretability and the PnP frameworks are very slow. 

Among the deep unfolding methods, Wang \emph{et al.} \cite{wang2019hyperspectral} unfolded the half-quadratic splitting (HQS) method and designed a spatial-spectral deep priors to boost the data fidelity. Furthermore, the local and non-local correlations of HSIs \cite{wang2020dnu} were considered in the prior design. Zhang \emph{et al.} \cite{zhang2021learning} learned the tensor low-rank spectral prior via the deep CP decomposition. Most recently, Huang \emph{et al.} \cite{huang2021deep} proposed a deep Gaussian scale mixture model to learn the scale prior and estimate the local means of images via the 3D filter. Although these methods have achieved great success, the lack of cross-phase feature interaction and content-aware parameter adjustment are still major bottlenecks for the reconstruction performance. Hereby, a novel DUN is proposed to effectively exploit the cross-phase correlation and update the parameters adaptively in this paper.

\subsection{Mask Optimization Algorithms}
Some existing works on traditional CS have explored the possibility of joint mask optimization and image reconstruction. For instance, Zhang \emph{et al.} \cite{zhang2020optimization} proposed a constrained optimization-inspired network for adaptive sampling and recovery. You \emph{et al.} \cite{you2021coast} introduced the random projection augmentation strategy to learn the arbitrary-sampling matrices and improve the generalization ability of the model. In the spectral SCI, Arguello \emph{et al.} \cite{arguello2012rank} transformed mask optimization into a rank minimization problem based on the theory of the Restricted Isometry Property (RIP) \cite{eldar2012compressed}. Furthermore, Wang \emph{et al.} \cite{wang2018hyperreconnet} rearranged the shifted 3D data cube and divided it into four parameter-sharing sub-patches for sampling mask learning. Simultaneously, Zhang \emph{et al.} \cite{zhang2021deep} designed an end-to-end learnable auto-encoder to optimize the illumination pattern and compress the HSIs. Although the above-mentioned methods realize the adaptive sampling to a certain extent, combining mask optimization with the DUN in the spectral SCI is still challenging and worth exploring.
\begin{figure*}[htbp]
	\centering
	\vspace{0.05cm}
	\includegraphics[width=1.\linewidth]{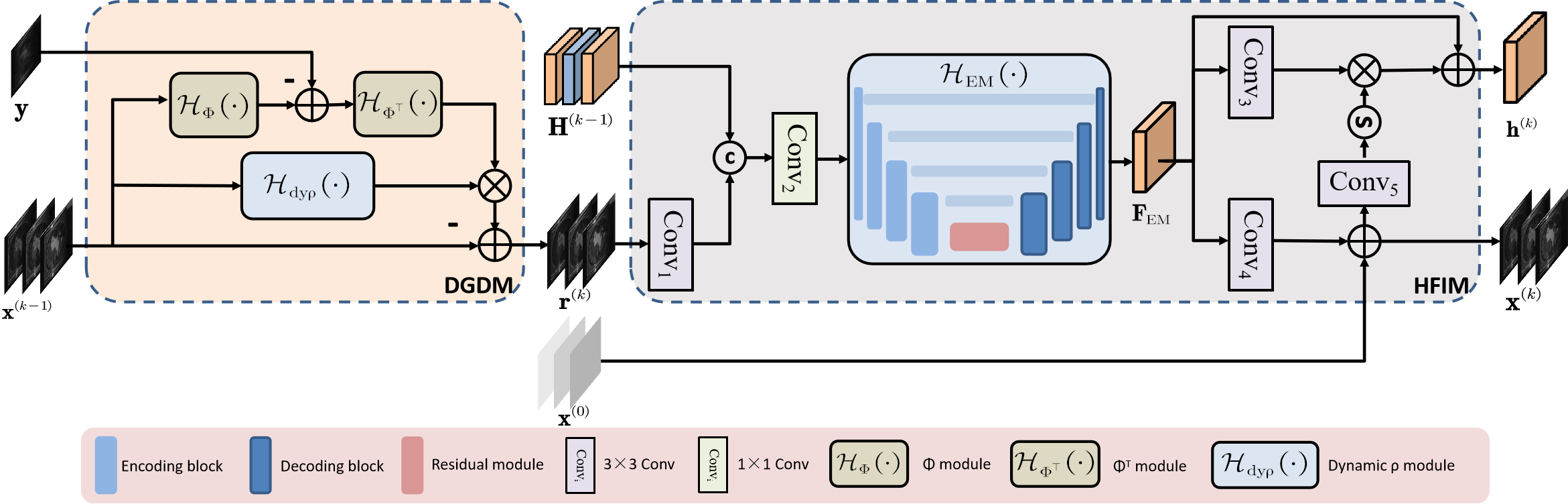}
	\caption{Details of the recovery phase in HerosNet. The recovery phase is composed of a dynamic gradient descent module (DGDM) and a hierarchical feature interaction module (HFIM). The DGDM takes the intermediate result $\mathbf{x}^{(k-1)}$ and the measurement $\mathbf{y}$ as input, and generates the coarse estimate $\mathbf{r}^{(k)}$. The HFIM is designed to refine the coarse estimate $\mathbf{r}^{(k)}$ with the hidden states of the previous phases $\mathbf{H}^{(k-1)}$ and the initialization cube $\mathbf{x}^{(0)}$ to produce the accurate reconstructed images $\mathbf{x}^{(k)}$ and the hidden states $\mathbf{h}^{(k)}$ in the $k^{th}$ phase.}
	\label{prior}
\end{figure*}
\section{Proposed Method}
\subsection{Problem Formulation}
\label{sec:cassi}
In CASSI system, the 3D hyperspectral cube is firstly modulated via a coded aperture (\emph{\emph{i.e.}} a physical mask) and then dispersed via a dispersive prism. Mathematically, considering a sequence $\{\mathbf{X}_{i}\}_{i=1}^{C} \in \mathbb{R}^{H \times W}$ composed of $C$ HSI frames, they are modulated via a mask $\mathbf{M} \in \mathbb{R}^{H \times W}$:
\begin{equation}
	\mathbf{X}_{i}^{\prime}=\mathbf{M} \odot \mathbf{X}_{i},
	\label{eq1}
\end{equation}
where $\mathbf{X}_{i}^{\prime}$ is the modulated HSI frame and $\odot$ is the Hadamard (element-wise) product. After that, modulated HSI frames with different wavelengths are shifted spatially and summed in an element-wise manner. Therefore, the modulated HSI frames $\{\mathbf{X}_{i}^{\prime}\}_{i=1}^{C} \in \mathbb{R}^{H \times W}$ are compressed to a coded measurement as follows:
\begin{equation}
	\mathbf{Y}(m,n)=\sum_{i=1}^{C} \mathbf{X}_{i}^{\prime}(m, n+d_i)+\mathbf{N},
	\label{eq2}
\end{equation}
where $m$, $n$ denote the spatial coordinates, and $d_i$ denotes the shifting distance of the $i^{th}$ channel. $\mathbf{N}$$\in $$\mathbb{R}^{H \times (W+C-1)}$ and $\mathbf{Y}$$\in$$\mathbb{R}^{H \times (W+C-1)}$ denote the noise and the compressed measurement, respectively. The vectorized form of the spectral SCI is expressed as follows:
\begin{equation}
	\mathbf{y}=\mathbf{\Phi} \mathbf{x}+\mathbf{n},
	\label{eq3}
\end{equation}
where $\mathbf{x}$$\in$$\mathbb{R}^{HWC}$, $\mathbf{y}$$\in$$\mathbb{R}^{H(W+C-1)}$, and $\mathbf{n}$$\in$$\mathbb{R}^{H(W+C-1)}$ denote the vectorized form of $\mathbf{X}$, $\mathbf{Y}$, and $\mathbf{N}$, respectively. $\mathbf{\Phi}\in\mathbb{R}^{H(W+C-1) \times HWC}$ represents the sensing matrix.

\subsection{Architecture of Proposed HerosNet}
In this subsection, we propose an optimization-inspired deep unfolding network for joint mask optimization and image reconstruction. As illustrated in Fig.~\ref{net}, the proposed HerosNet is composed of three subnets, including a sampling subnet, an initialization subnet, and a recovery subnet. 

\subsubsection{Sampling Subnet}
In this paper, the sampling subnet aims to learn the optimal binary mask for the HSI compressive sensing, which preserves enough spectral-spatial information and eliminates redundancy. The training process of the sampling subnet is divided into three stages, including randomization, binarization, and compression. To learn the binary mask $\mathbf{M}$, random Gaussian initialization with the mean $\mu_b$ and the variance $\sigma_b$ is adopted to generate a continuous matrix $\tilde{\mathbf{M}}$. Furthermore, we design an element-wise binarization function $\operatorname{BinarySign(\cdot)}$ to convert the continuous matrix into the binary mask as follows:
\begin{equation}
	\mathbf{M}=\operatorname{BinarySign}(\tilde{\mathbf{M}}),
	\label{eq4}
\end{equation}  
\begin{equation}
with~~~~ \operatorname{BinarySign}(z)= 1  \text { if } z \geq \mu_b  \text{\  or \ }  0  \text {\ else}.
	\label{eq5}
\end{equation}
According to the imaging rule depicted in the Sec.~\ref{sec:cassi}, we utilize a conversion function $\operatorname{Mask2Mat(\cdot)}$ to transform the binary mask $\mathbf{M}$ into the sensing matrix $\mathbf{\Phi}$ in Eq.~(\ref{eq3}):
\begin{equation}
   \mathbf{\Phi}=\operatorname{Mask2Mat}(\mathbf{M}).
\end{equation}
Since the sensing matrix $\mathbf{\Phi}$ is regarded as the learnable parameter, the derivative of the binarization function is defined as a constant, i.e. $\operatorname{BinarySign}^{\prime}(z)=1$, for the backpropagation of the sampling subnet. Finally, according to Eq.~(\ref{eq1}) and Eq.~(\ref{eq2}), the 3D hyperspectral cube $\mathbf{x}$ is compressed to the snapshot measurement $\mathbf{y}$.

\subsubsection{Initialization Subnet}
Given the measurement $\mathbf{y} \in \mathbb{R}^{H \times (W+C-1)}$, the initialization subnet aims to split this 2D measurement into the 3D hyperspectral cube $\mathbf{x}^{(0)} \in \mathbb{R}^{H \times W \times C}$, where $W$, $H$ are the spatial size of the frames, and $C$ is the number of spectral channels. Specifically, an extraction window is cropped from the measurement $\mathbf{y}$ and then slided in the step size of $d$ to generate $C$ HSI frames. Finally, the $C$ frames are concatenated in the channel dimension to compose a 3D hyperspectral cube $\mathbf{x}^{(0)} \in \mathbb{R}^{H \times W \times C}$.

\subsubsection{Recovery Subnet}
As depicted in Fig.~\ref{prior}, the proposed recovery subnet aims to reconstruct high-quality HSIs from the compressive measurement. Inspired by the ISTA, the image reconstruction is treated as an optimization problem shown as follows:
\begin{equation}
	\mathbf{\mathbf{x}}=\arg \min _{\mathbf{x}} \frac{1}{2}\|\mathbf{y}-\mathbf{\Phi} \mathbf{x}\|_{2}^{2}+\lambda \psi(\mathbf{x}).
	\label{eq6}
\end{equation}
To be noted, the first term is the data fidelity term, while the second term $\psi(\cdot)$ is the prior regularization term. $\lambda$ denotes a regularization parameter.

To solve the optimization problem in Eq.~(\ref{eq6}), we unfold ISTA to design the deep network for its simplicity and interpretability. Traditional ISTA updates the results via two steps, namely gradient descent and proximal mapping, which are formulated as follows:
\begin{equation}
	\mathbf{r}^{(k)}=\mathbf{x}^{(k-1)}-\rho \mathbf{\Phi}^{\top}(\mathbf{\Phi} \mathbf{x}^{(k-1)}-\mathbf{y}),
	\label{eq7}
\end{equation}
\begin{equation}
	\mathbf{x}^{(k)}=\underset{\mathbf{x}}{\arg \min } \frac{1}{2}\|\mathbf{x}-\mathbf{r}^{(k)}\|_{2}^{2}+\lambda \psi(\mathbf{x}),
	\label{eq8}
\end{equation}
where $k$ denotes the number of ISTA iteraction and $\rho$ denotes the step size. By introducing the proximal mapping operator $\operatorname{prox}_{\lambda \psi}(\mathbf{r})=\arg \min _{\mathbf{x}} \frac{1}{2}\|\mathbf{x}-\mathbf{r}\|_{2}^{2}+\lambda \psi(\mathbf{x})$, Eq.~(\ref{eq8}) can be rewritten as follows:
\begin{equation}
	\mathbf{x}^{(k)}=\operatorname{prox}_{\lambda \psi}(\mathbf{r}^{(k)}).
	\label{eq9}
\end{equation}
Modifying these two steps, we design a dynamic gradient descent module (DGDM) and a hierarchical feature interaction module (HFIM) to reconstruct the HSIs.

\begin{figure}[!]
	\centering
	\includegraphics[width=1\linewidth]{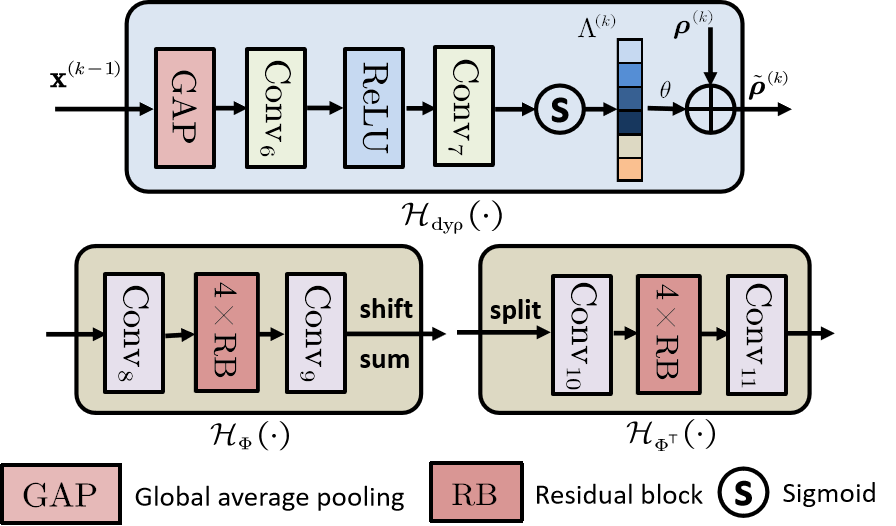}
	\caption{Details of some key components in the DGDM. The DGDM aims to simulate the sensing matrix flexibly and adjust the step size dynamically.}
	\label{dyrou}
\end{figure}

\noindent \textbf{Dynamic Gradient Descent Module (DGDM):} To implement Eq.~(\ref{eq7}) via the deep network, the DGDM is employed to generate the immediate reconstructed results $\mathbf{r}^k$ dynamically. Most existing DUNs used to treat $\mathbf{\Phi}$, $\mathbf{\Phi}^{\top}$ and $\rho$ in Eq.~(\ref{eq7}) as the constants, which restricts the flexibility of the network and limits the accuracy of reconstruction. To address these issues, the deep modules $\mathcal{H}_{\mathbf{\Phi}}(\cdot)$ and $\mathcal{H}_{\mathbf{\Phi}^{\top}}(\cdot)$ are introduced to simulate the matrix $\mathbf{\Phi}$ and $\mathbf{\Phi}^{\top}$ from the previous state $\mathbf{x}^{(k-1)}$, where $\mathcal{H}_{\mathbf{\Phi}}(\cdot)$ and $\mathcal{H}_{\mathbf{\Phi}^{\top}}(\cdot)$ consist of two convolution operators and four residual blocks, respectively. In order to achieve content-aware parameter adjustment, a dynamic step size operator $\mathcal{H}_{\text{dy}\rho}(\cdot)$ is incorporated into the process of gradient descent to further enhance the generalization ability of the network. The step size $\boldsymbol{\tilde{\rho}}^{(k)}$ is directly learned from the previous state $\mathbf{x}^{(k-1)}$ and adjusted adaptively with the advance of network training. Specifically, we decompose $\boldsymbol{\tilde{\rho}}^{(k)}$ into the static and dynamic components. The static component is a learnable vector and the values of each spectral channels are weight-sharing. The dynamic component is a channel attention map learned from $\mathbf{x}^{(k-1)}$. As illustrated in Fig.~\ref{dyrou}, the channel attention map is obtained by the global average pooling, two $1$$\times$$1$ convolution operators, ReLU activation function, and the Sigmoid function. Finally, Eq.~(\ref{eq7}) can be modified as follows:
\begin{equation}
	\begin{aligned}
		\mathbf{r}^{(k)} &= \operatorname{DGDM} (\mathbf{x}^{(k-1)}, \mathbf{y}) \\
		&=\mathbf{x}^{(k-1)}-\boldsymbol{\tilde{\rho}}^{(k)}\mathcal{H}_{\mathbf{\Phi}^{\top}}(\mathcal{H}_{\mathbf{\Phi}}(\mathbf{x}^{(k-1)})-\mathbf{y}),
		\label{eq10}
	\end{aligned}
\end{equation}
\begin{equation}
with~~~~	\boldsymbol{\tilde{\rho}}^{(k)}=\mathcal{H}_{\text{dy}\rho}(\mathbf{x}^{(k-1)})=\boldsymbol{\rho}^{(k)}+\theta \boldsymbol{\Lambda^{(k)}},
	\label{eq11}
\end{equation}
where $\boldsymbol{\rho}^{(k)}$$\in$$\mathbb{R}^{1 \times C}$ and $\boldsymbol{\Lambda}^{(k)}$$\in$$\mathbb{R}^{1 \times C}$ denote the static and the dynamic component, respectively. $\theta$ is a constant to stabilize the network training.

\noindent \textbf{Hierarchical Feature Interaction Module (HFIM):} To implement Eq.~(\ref{eq8}) via the deep network, the HFIM is designed to refine the coarse reconstructed result $\mathbf{r}^{(k)}$. There are two issues restricting the performance of previous deep proximal mapping modules. Firstly, since the gradient descent module functions in the image domain and the deep proximal mapping module performs in the feature domain, the spectral information will be lost when it is transmitted between these two modules. Secondly, as the number of recovery phases increases, the useful features in the previous phases are not propagated to the subsequent phases. To alleviate these issues, the proposed HFIM focuses on two aspects: 1) how to bridge the gap between the image domain and the feature domain; 2) how to effectively interact the beneficial features in previous phases to enhance the features of the current phase.

As illustrated in Fig.~\ref{prior}, in addition to $\mathbf{r}^{(k)}$ produced from the DGDM, the HFIM takes the cascaded hierarchical features $\mathbf{H}^{(k-1)}=[\mathbf{h}^{(k-1)}, \mathbf{h}^{(k-2)}, \ldots, \mathbf{h}^{(0)}]$ as input, where $\mathbf{h}^{(i)}$ denotes the hidden state of the $i^{th}$ phase. To be noted, the hidden states retain the beneficial information extracted from the reconstructed image in the current phase. Concretely, the intermediate result $\mathbf{r}^{(k)}$ is firstly transformed into the feature domain via the operator $\text{Conv}_{1}$ and fused with hidden states $\mathbf{H}^{(k-1)}$ via the dense connection $\text{Conv}_{2}$. Then, the fused feature is fed to an enhancement module $\mathcal{H}_{\text{EM}}(\cdot)$ to extract the spectral and spatial features, where $\mathcal{H}_{\text{EM}}(\cdot)$ is composed of four encoding blocks and four decoding blocks. To enhance the representation ability of the network, the residual module consisting of sixteen standard residual blocks is embedded between the encoder and decoder. The process is formulated as follows:
\begin{equation}
    \mathbf{F}_{\text{EM}} = \mathcal{H}_{\text{EM}}(\text{Conv}_{2}([\mathbf{H}^{(k-1)}, \text{Conv}_{1}(\mathbf{r}^{(k)})])),
	\label{eq12}
\end{equation}
where $\mathbf{F}_{\text{EM}}\in \mathbb{R}^{H \times W \times N}$ denotes the enhanced feature produced by $\mathcal{H}_{\text{EM}} (\cdot)$, $N$ denotes the channel numbers.

After obtaining the enhanced feature $\mathbf{F}_{\text{EM}} \in \mathbb{R}^{H \times W \times N}$, a well-designed feature interaction mechanism is designed to generate the reconstructed HSI frames $\mathbf{x}^{(k)}\in\mathbb{R}^{H \times W \times C}$ and the hidden state $\mathbf{h}^{(k)}\in\mathbb{R}^{H \times W \times N}$. On the one hand, the enhanced feature $\mathbf{F}_{\text{EM}}$ is transformed into the image domain via the operator $\text{Conv}_{4}$ and added with the initialization cube $\mathbf{x}^{(0)}$ to generate the reconstructed HSI frames $\mathbf{x}^{(k)}$. The process is exhibited as follows:
\begin{equation}
    \mathbf{x}^{(k)} = \text{Conv}_{4}(\mathbf{F}_{\text{EM}})+\mathbf{x}^{(0)}.
	\label{eq13}
\end{equation}
On the other hand, the network directly learns an attention cube from the current state $\mathbf{x}^{(k)}$ via the operators $\text{Conv}_5$ and $\text{Sigmoid}$ to provide pixel-level interactive information and generate the hidden state $\mathbf{h}^{(k)}$, which is formulated as:
\begin{equation}
    \mathbf{h}^{(k)} = \text{Conv}_{3}(\mathbf{F}_{\text{EM}}) \otimes \text{Sigmoid}(\text{Conv}_{5}(\mathbf{x}^{(k)})) + \mathbf{F}_{\text{EM}}.
	\label{eq14}
\end{equation}
Both $\mathbf{x}^{(k)}$ and $\mathbf{h}^{(k)}$ will be utilized in the reconstruction of subsequent phases. Finally, Eq.~(\ref{eq8}) can be modified as:
\begin{equation}
	\mathbf{x}^{(k)}, \mathbf{h}^{(k)}= \operatorname{HFIM} (\mathbf{r}^{(k)}, \mathbf{x}^{(0)}, \mathbf{H}^{(k-1)}).
	\label{eq15}
\end{equation}

By introducing the HFIM, we establish the hierarchical connection to integrate features in different phases. With the proposed DGDM and HFIM, the recovery subnet can reconstruct HSIs accurately and rapidly.  

\begin{table*}[]
	\renewcommand{\arraystretch}{1.}
	\centering
	\caption{Comparison results of the proposed network and state-of-the-art HSI reconstruction methods on the KAIST dataset. To be noted, \emph{HerosNet-base} denotes the proposed method without mask optimization. Best results are in {\color{red}{red}} and the second-best ones are in {\color{blue}{blue}}. }
	\centering
	\label{table1}
	\resizebox{\textwidth}{!}{
	\begin{tabular}{c|ccccccc|cc}
        \toprule[1.5pt]
		\multirow{2}{*}{\textbf{Testing Set}} & \textbf{GAP-TV \cite{yuan2016generalized}} & \textbf{DeSCI \cite{liu2018rank}} & \textbf{HSSP \cite{wang2019hyperspectral}} & \textbf{$\lambda$-net \cite{miao2019net}} & \textbf{TSA-Net \cite{meng2020end}} & \textbf{PnP-DIP-HSI \cite{meng2021self}} & \textbf{DGSM \cite{huang2021deep}} & \textbf{HerosNet-base} & \textbf{HerosNet}      \\
		& (ICIP, 2016) & (TPAMI, 2018)  & (CVPR, 2019)   & (ICCV, 2019)  & (ECCV, 2020) & (ICCV, 2021) & (CVPR, 2021) & (Ours) & (Ours)                 \\ \midrule[0.75pt]
		Scene01           & 25.13 / 0.724  & 27.15 / 0.794  & 31.07 / 0.852  & 30.82 / 0.880 & 31.26 / 0.887 & 32.70 / 0.898 & 33.17 / 0.954 & {\color{blue}{34.24}} / {\color{blue}{0.963}} & \textbf{\color{red}{35.69}} / \textbf{\color{red}{0.973}} \\
		Scene02           & 20.67 / 0.630  & 22.26 / 0.694  & 26.30 / 0.798  & 26.30 / 0.846 & 26.88 / 0.855 & 27.27 / 0.832 & 31.61 / 0.933 & {\color{blue}{32.94}} / {\color{blue}{0.952}} & \textbf{\color{red}{35.01}} / \textbf{\color{red}{0.968}} \\
		Scene03           & 23.19 / 0.757  & 26.56 / 0.877  & 29.00 / 0.875  & 29.42 / 0.916 & 30.03 / 0.921 & 31.32 / 0.920 & 31.55 / 0.952 & {\color{blue}{34.15}} / {\color{blue}{0.966}} & \textbf{\color{red}{34.82}} / \textbf{\color{red}{0.967}} \\
		Scene04           & 35.13 / 0.870  & 39.00 / 0.965  & 38.24 / 0.926  & 37.37 / 0.962 & {\color{blue}{39.90}} / 0.964 & \textbf{\color{red}{40.79}} / 0.970 & 37.43 / 0.981 & 38.80 / \color{blue}{0.984} & 38.07 / \textbf{\color{red}{0.985}} \\
		Scene05           & 22.31 / 0.674  & 24.80 / 0.778  & 27.98 / 0.827  & 27.84 / 0.866 & 28.89 / 0.878 & 29.81 / 0.903 & 29.43 / 0.927 & {\color{blue}{31.39}} / {\color{blue}{0.953}} & \textbf{\color{red}{33.18}} / \textbf{\color{red}{0.969}} \\
		Scene06           & 22.90 / 0.635  & 23.55 / 0.753  & 29.16 / 0.823  & 30.69 / 0.886 & 31.30 / 0.895 & 30.41 / 0.890 & 32.49 / 0.960 & {\color{blue}{32.88}} / {\color{blue}{0.960}} & \textbf{\color{red}{34.94}} / \textbf{\color{red}{0.976}} \\
		Scene07           & 17.98 / 0.670  & 20.03 / 0.772  & 24.11 / 0.851  & 24.20 / 0.875 & 25.16 / 0.887 & 28.18 / 0.913 & 30.64 / 0.937 & {\color{blue}{32.79}} / \textbf{\color{red}{0.963}} & \textbf{\color{red}{33.58}} / \color{blue}{0.962} \\
		Scene08           & 23.00 / 0.624  & 20.29 / 0.740  & 27.94 / 0.831  & 28.86 / 0.880 & 29.69 / 0,887 & 29.45 / 0.885 & 31.06 / {\color{blue}{0.955}} & {\color{blue}{31.11}} / 0.953 & \textbf{\color{red}{33.19}} / \textbf{\color{red}{0.968}} \\
		Scene09           & 23.36 / 0.717  & 23.98 / 0.818  & 29.14 / 0.822  & 29.32 / 0.902 & 30.03 / 0.903 & \textbf{{\color{red}{34.55}}} / 0.932 & 30.87 / 0.951 & 31.58 / \color{blue}{0.953} & {\color{blue}{33.04}} / \textbf{\color{red}{0.964}} \\
		Scene10           & 23.70 / 0.551  & 25.94 / 0.666  & 26.44 / 0.740  & 27.66 / 0.843 & 28.32 / 0.848 & 28.52 / 0.863 & 31.34 / {\color{blue}{0.955}} & {\color{blue}{31.64}} / 0.949 & \textbf{\color{red}{33.01}} / \textbf{\color{red}{0.965}} \\ \midrule[0.75pt]
		Average           & 23.73 / 0.683  & 25.86 / 0.785  & 28.93 / 0.834  & 29.25 / 0.886 & 30.15 / 0.893 & 31.30 / 0.901 & 31.96 / 0.951 & {\color{blue}{33.15}} / {\color{blue}{0.960}} & \textbf{\color{red}{34.45}} / \textbf{\color{red}{0.970}} \\ \bottomrule[1.5pt]
	\end{tabular}}
\end{table*}

\subsection{Network Training and Implementation Details}
In our implementation, the number of spectral channels $C$ is 28 and the number of feature channels $N$ is 32. In our sampling subnet with mask optimization, $\mu_b$ and $\sigma_b$ are set to 0 and 0.1, respectively. In our recovery subnet, the number of recovery phases $K$ is 8 and $\theta$ in Eq.~(\ref{eq11}) is 0.5. The learnable parameters in the proposed network are denoted by $\mathbf{\Theta}$, including the binary mask $\mathbf{M}$, the parameters $\mathbb{D}^{(k)}$ in the DGDM and $\mathbb{H}^{(k)}$ in the HFIM. To learn the parameters $\mathbf{\Theta}=\{\mathbf{M}, \mathbb{D}^{(k)}, \mathbb{H}^{(k)}\}$, we utilize the reconstructed results of the final phase and some intermediate phases to calculate the loss function \cite{meng2020gap}. Specifically, given the training data $\{\mathbf{x}_{i}\}_{i=1}^{N_{d}}$, the loss function is defined:
\begin{equation}
	\mathcal{L}=\mathcal{L}_{f}+\mathcal{L}_{p},
	\label{eq16}
\end{equation}
\begin{equation}
with~~~~	\mathcal{L}_{f}=\frac{1}{N_{d}} \sum_{i=1}^{N_{d}}\|\mathbf{x}_{i}^{(K)}-\mathbf{x}_{i}\|_{2}^{2},
	\label{eq17}
\end{equation}
\begin{equation}
	\mathcal{L}_{p}=\frac{\beta}{N_{d}} \sum_{i=1}^{N_{d}}\|\mathbf{x}_{i}^{(K-1)}-\mathbf{x}_{i}\|_{2}^{2}+\|\mathbf{x}_{i}^{(K-2)}-\mathbf{x}_{i}\|_{2}^{2}.
	\label{eq18}
\end{equation}
where $\mathcal{L}_{f}$ and $\mathcal{L}_{p}$ denote the loss function of the final phase and the previous few phases, respectively. $K$ and $N_{d}$ denote the number of recovery phases and training samples. $\beta$ is the equilibrium constant and set to 0.5.

Our HerosNet is implemented with 4 NVIDIA Tesla V100 GPUs. We adopt Adam \cite{kingma2014adam} to train the network for 100 epochs. The learning rate is initialized with $1$$\times$$10^{-4}$ and decays with a factor of 0.9 every 10 epochs.

\section{Experimental Results}
\subsection{Experimental Settings}
In this paper, we have verified the effectiveness of the proposed network on both simulation datasets and the real dataset. Following the settings of TSA-Net \cite{meng2020end} and DGSM \cite{huang2021deep}, the simulation experiments are conducted on the public HSI datasets CAVE \cite{yasuma2010generalized} and KAIST \cite{choi2017high} with the size $256$$\times$$256$$\times$$28$, i.e., 28 spectral channels with the spatial size $256$$\times$$256$. For the experiments in the real scenes, 5 compressive measurements with the spatial size of $640$$\times$$694$ captured by the real SCI system  are utilized for testing. The metrics of PSNR and SSIM \cite{wang2004image} are employed to evaluate the reconstruction quality.  
\begin{figure}[!]
	\centering
	\includegraphics[width=1\linewidth]{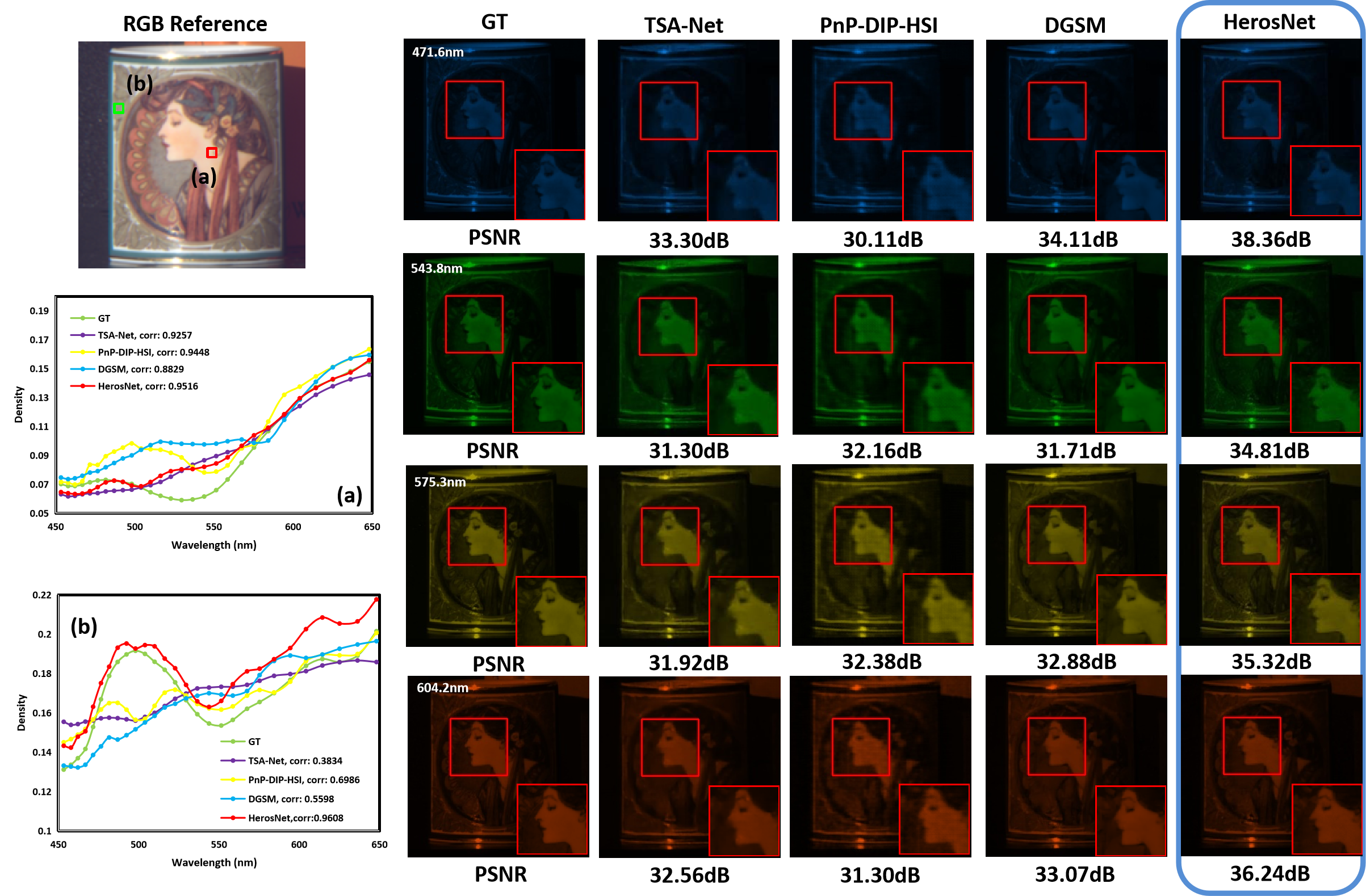}
	\caption{Reconstructed images of \emph{Scene1} on KAIST dataset by our HerosNet and other state-of-the-art methods. The HSI frames are converted to RGB images via the CIE color matching function \cite{smith1931cie}. Spectral curves on the selected regions ((a) and (b)) and the visualization results show that our results have higher spectral accuracy and better perceptual quality.}
	\label{sota1}
\end{figure}

\begin{figure}[htbp]
	\centering
	\includegraphics[width=1\linewidth]{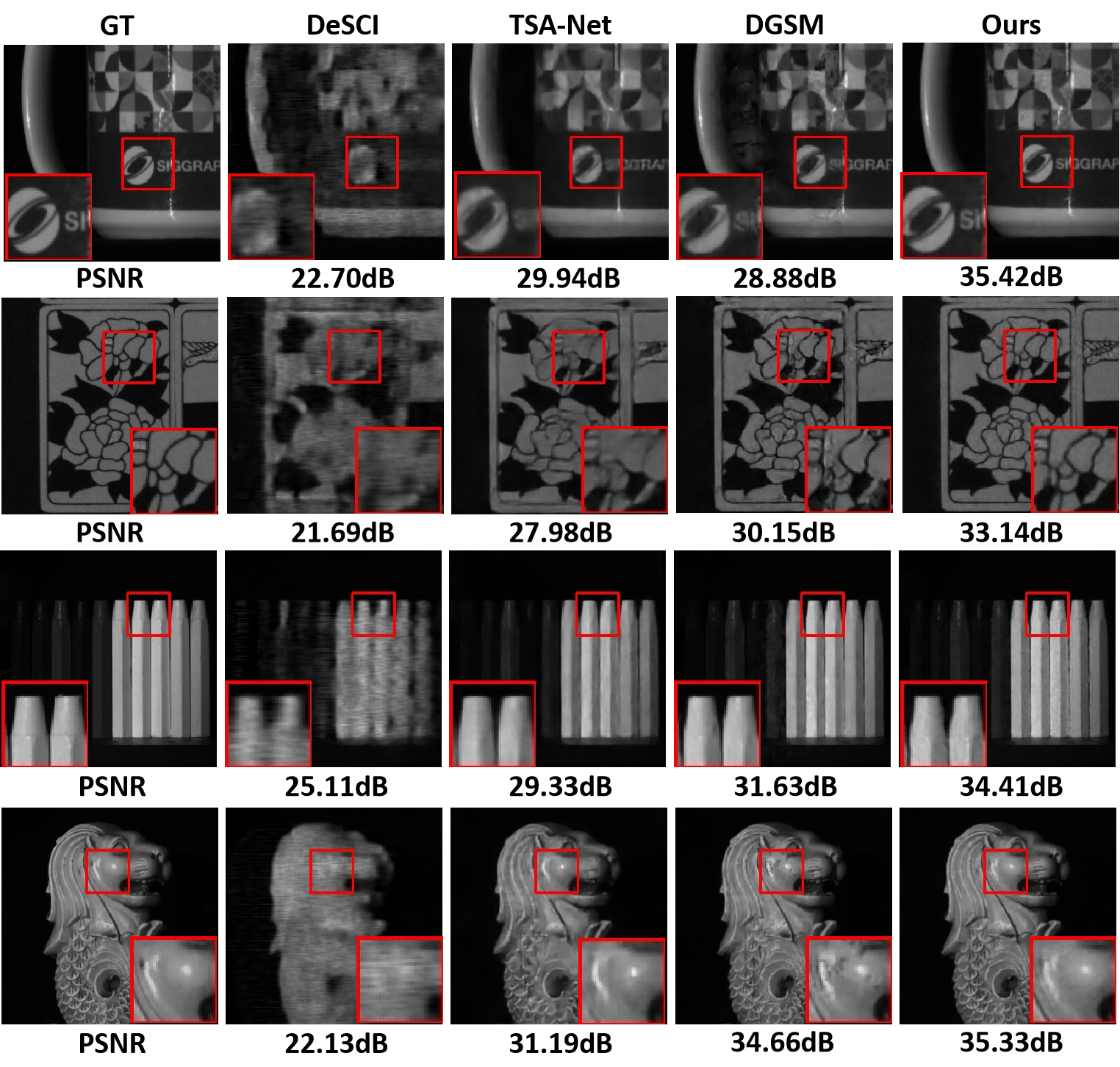}
	\caption{Visual comparisons of our HerosNet and other state-of-the-art methods on the KAIST dataset. The wavelengths of all these images are 648.1nm. The reconstructed images of HerosNet can preserve more details and clearer textures with less artifacts.}
	\label{sota}
\end{figure}

\subsection{Simulation Results}
To demonstrate the effectiveness of the proposed method on the simulation datasets, we compare the proposed HerosNet with several existing methods, including the model-based methods (GAP-TV \cite{yuan2016generalized}, DeSCI \cite{liu2018rank}), the end-to-end deep learning-based methods ($\lambda$-net \cite{miao2019net} and TSA-Net \cite{meng2020end}), the plug-and-play method (PnP-DIP-HSI \cite{meng2021self}) and the deep unfolding networks (HSSP \cite{wang2019hyperspectral} and DGSM \cite{huang2021deep}). All of these competing methods are trained on the CAVE dataset with a fixed real mask \cite{huang2021deep}.

As shown in Table~\ref{table1}, the proposed HerosNet obtains 34.45 dB of PSNR and 0.970 of SSIM, which has surpassed all of competing methods by large margins. Compared with the state-of-the-art method DGSM, the proposed network achieves 2.49dB improvement of PSNR and 0.019 improvement of SSIM. In comparison with the second best method PnP-DIP-HSI, the proposed method also achieves 3.15dB/0.069 gains on PSNR/SSIM. It demonstrates that the proposed recovery subnet can better unearth the spectral information and the proposed mask optimization strategy can search for the optimal binary mask. As illustrated in Fig.~\ref{sota1}, the reconstructed HSIs produced by the HerosNet have more spatial details and clearer texture in different spectral channels, while the results of other competing methods are blurry. In addition, the spectral curves of the HerosNet have a higher correlation with the reference spectra. Fig.~\ref{sota} further plots some visual comparisons of the proposed HerosNet, DGSM, TSA-Net and DeSCI on four other scenes. Compared with these three typical methods, the proposed HerosNet provides sharper edges, better visual effects and less artifacts. 
\begin{table}[]
	\renewcommand{\arraystretch}{1.}
	\centering
	\caption{Comparison results of the proposed optimized mask and other kinds of fixed masks. To be noted, even training on the same real mask \cite{huang2021deep}, the proposed method has surpassed all of the existing methods listed in Table~\ref{table1}.}
	\label{table2}
	\resizebox{!}{1.1cm}{
		\begin{tabular}{c|c|c}
			\toprule[1.5pt]
			Mask Type & PSNR  & SSIM  \\ \midrule[0.75pt]
			Uniform mask & 31.78 & 0.935 \\
			Guassian mask & 32.49 & 0.943 \\
			Real mask \cite{huang2021deep} & {\color{blue}{33.15}} & {\color{blue}{0.960}} \\
			Optimized binary mask & \textbf{\color{red}{34.45}} & \textbf{\color{red}{0.970}} \\ \bottomrule[1.5pt]
	\end{tabular}
}
\end{table}
\begin{figure}[!]
	\centering
	\includegraphics[width=1\linewidth]{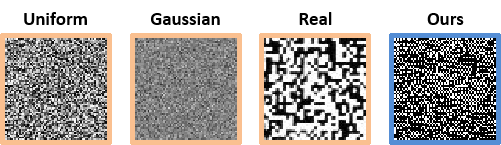}
	\caption{Visual illustrations of different kinds of masks, including the uniform mask, Gaussian mask, real fixed mask, optimized binary mask (Ours).}
	\label{mask}
\end{figure}

\subsection{Multiple Mask Results}
To objectively evaluate the effectiveness of the proposed mask optimization strategy, we train our model on different kinds of masks. As shown in Fig.~\ref{mask}, the center areas of four different masks with the spatial size $64$$\times$$64$ in the experiments are presented. To be noted, we remove the mask optimization strategy from the proposed network when training on the fixed mask. Table~\ref{table2} lists the PSNR and SSIM results by testing on the KAIST dataset. It can be clearly seen that the proposed HerosNet with joint mask optimization outperforms the networks that are directly trained on the fixed masks. The main reason is that the optimized binary mask preserves the complete image structure and sufficient detailed information to achieve optimal sampling. Simultaneously, the proposed method trained on the real mask also surpasses all the SOTA methods listed in Table~\ref{table1}. It further proves that even training on the same mask, the proposed method also has great advantages in HSI reconstruction.

\begin{figure}[!]
	\centering
	\includegraphics[width=1.\linewidth]{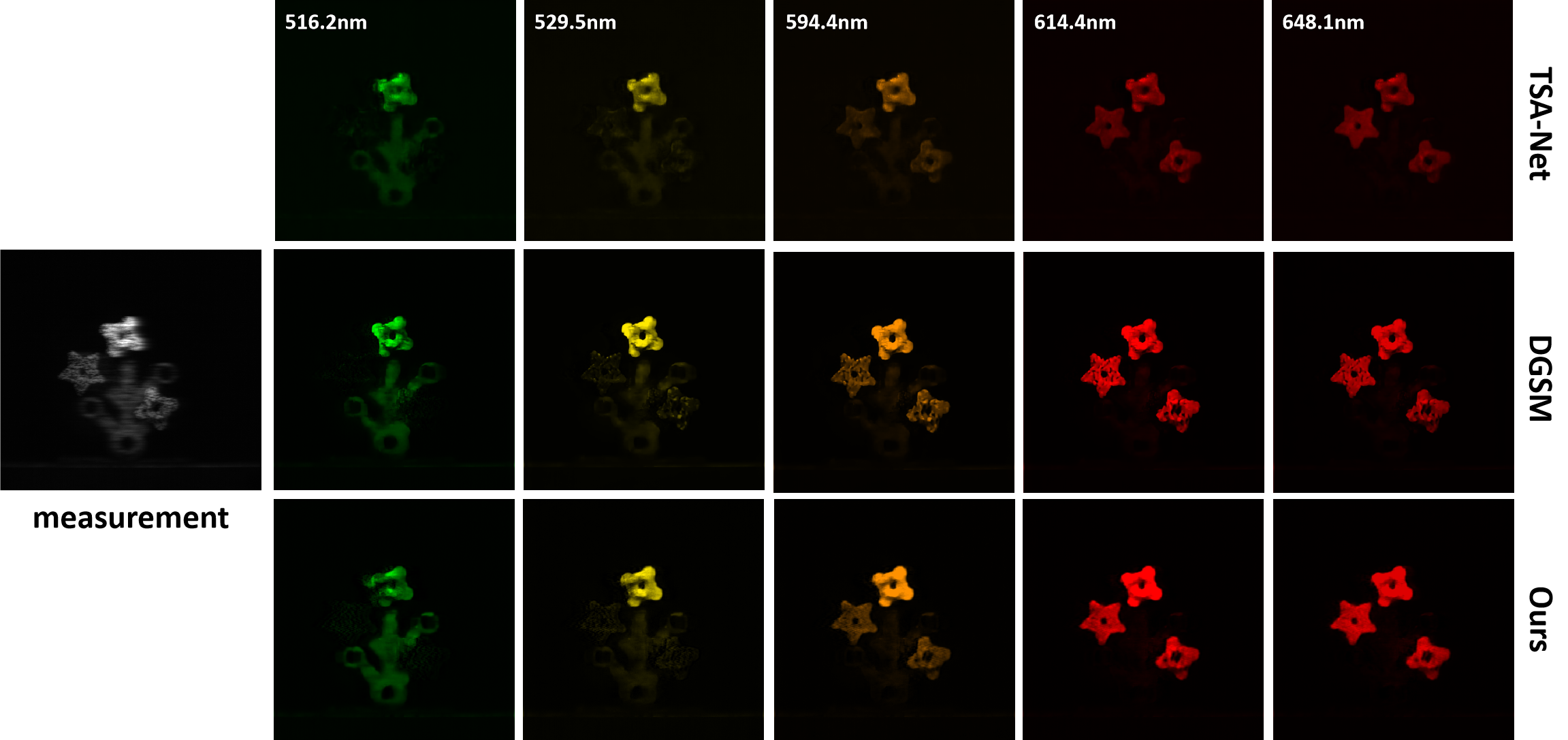}
	\caption{Visual comparisons of our HerosNet, DGSM \cite{huang2021deep} and TSA-Net \cite{meng2020end} on the real dataset \emph{Scene1}. Obviously, our HerosNet can recover more image details and clearer content (i.e. the flower in the right corner) than DGSM and TSA-Net.}
	\label{real}
\end{figure}

\subsection{Real Data Results}
To verify the effect of the proposed method on the real data, five compressive measurements captured by the real spectral SCI system are utilized for testing. Due to the ground truths of real scenes are unavailable, we get rid of the sampling subnet and only take the 2D compressive measurements as input. For fair comparisons, all of the methods are trained on the CAVE datasets using the fixed real mask with 11-bit shot noise injected. Fig.~\ref{real} plots the visual comparisons of the proposed HerosNet and the existing SOTA method DGSM \cite{huang2021deep}, TSA-Net \cite{meng2020end}. Obviously, our results recover more image details with fewer artifacts and clearer content in various wavelengths, while the reconstructed images of other methods are blurrier and more susceptible to noise corruption. It further proves that the proposed HerosNet is more robust to the noise distortion and effective in the real spectral imaging system. 

\subsection{Ablation study}
To evaluate the contribution of different components in the proposed HerosNet, ablation study is conducted on the CAVE and KAIST datasets. We mainly focus on the four components, namely, mask optimization (MO), hierarchical feature interaction module (HFIM), residual module (RM), and dynamic step size (Dy$\rho$) mechanism. Table~\ref{table3} shows the results of PSNR and SSIM on different settings. To investigate the validity of mask optimization strategy, we remove the MO and retrain our reconstruction network on the fixed real mask. It can be clearly seen that the PSNR and SSIM results have a decline by 1.30dB and 0.010 respectively, which proves the effectiveness of the proposed MO. To be noted, even without the MO, \emph{Ours-base} also achieves the best performance among all the existing reconstruction methods listed in Table~\ref{table1}. To investigate the impact of HFIM, we reimplement a variant network, which directly utilizes a U-Net as the deep prior without any interaction between phases. Obviously, without HFIM, the values of PSNR and SSIM have dropped by 1.04dB and 0.009 respectively, thus proving its significant effect. Meanwhile, Table~\ref{table3} shows a substantial drop on PSNR/SSIM from 34.45dB/0.970 to 33.59dB/0.965 when RM is removed. Furthermore, replacing the dynamic step size $\boldsymbol{\tilde{\rho}}^{(k)}$ with the static component $\boldsymbol{\rho}^{(k)}$ in Eq.~(\ref{eq11}), the results of PSNR and SSIM have decreased by 0.34dB and 0.004 respectively, which verifies the role of dynamic step size mechanism. 
\begin{table}[!]
	\renewcommand{\arraystretch}{1.}
	\centering
	\caption{Evaluation of the effectiveness of different components.}
	\label{table3}
	\resizebox{!}{1.2cm}{
		\centering
		\begin{tabular}{c|c|c|c|c|c|c}
			\toprule[1.5pt]
			Case Index & MO & HFIM & RM & Dy$\rho$ & PSNR  & SSIM  \\ \midrule[0.75pt]
			(a) (Ours-base) & $\times$   & $\checkmark$     & $\checkmark$  & $\checkmark$   & 33.15 & 0.960 \\
			(b)  & $\checkmark$  & $\times$       & $\checkmark$  & $\checkmark$   & 33.41 & 0.961 \\
			(c)  & $\checkmark$  & $\checkmark$     & $\times$    & $\checkmark$   & 33.59 & 0.965 \\
			(d)  & $\checkmark$  & $\checkmark$     & $\checkmark$  & $\times$     & 34.11 & 0.966 \\
			(e) (Ours) & $\checkmark$  & $\checkmark$     & $\checkmark$  & $\checkmark$   & 34.45 & 0.970 \\ \bottomrule[1.5pt]
	\end{tabular}}
\end{table}
\section{Conclusion}
In this paper, we propose a novel HerosNet for spectral snapshot compressive imaging. Inspired by the ISTA, HerosNet unfolds the optimization iterative process and is capable of jointly optimizing binary masks and reconstructing the HSIs accurately. To improve the generalization ability and flexibility of the network, a dynamic gradient descent module is proposed to achieve adaptive and content-aware parameter adjustment. To better utilize the cross-phase correlation, a hierarchical feature interaction module is designed to fuse and interact the useful information between different phases. Finally, experiments demonstrate that our network outperforms the state-of-the-art methods on both simulation and real datasets. Our future work will support HerosNet on MindSpore~\cite{mindspore}, which is a new deep learning computing framework.

\noindent\textbf{Broader impacts and limitations:} The proposed HerosNet contributes to the industrial application of spectral SCI and inspires the design of deep unfolding networks in other image inverse problems. Whereas, our model can not obtain decent results without retraining or fine-tuning when it comes to different imaging systems and physical masks. Meanwhile, the proposed learning-based method will inevitably reflect biases in the training data. These issues warrant further research and exploration for application.



{\small
\bibliographystyle{ieee_fullname}
\bibliography{egbib}
}

\end{document}